\documentclass[screen]{acmart}

\AtBeginDocument{%
  \providecommand\BibTeX{{%
    \normalfont B\kern-0.5em{\scshape i\kern-0.25em b}\kern-0.8em\TeX}}}

\setcopyright{acmcopyright}
\copyrightyear{2022}
\acmYear{2022}
\acmDOI{XXXXXXX.XXXXXXX}

\acmConference[Conference acronym 'XX]{Make sure to enter the correct
  conference title from your rights confirmation emai}{June 03--05,
  2018}{Woodstock, NY}
\acmPrice{15.00}
\acmISBN{978-1-4503-XXXX-X/18/06}

\usepackage{xcolor}
\usepackage{xspace}
\usepackage{soul}

\def\eg{\emph{e.g.}\xspace}

\newcommand{\apy}[1]{{#1}}

\newcommand{\pair}[1]{\left({#1}\right)}

\newcommand{\set}[1]{\left\{{#1}\right\}}
\newcommand{\ang}[1]{\left\langle{#1}\right\rangle}

\newcommand{\nat}{\mathbb{N}}

\newcommand{\real}{\mathbb{R}}

\usepackage{tikz} 
\usetikzlibrary{graphs}
\usetikzlibrary{arrows}
\usetikzlibrary{arrows,positioning,automata,calc}

\begin{document}

\title{A Graph-Based Context-Aware Model to Understand Online Conversations}


\author{Vibhor Agarwal}
\authornote{Author emails: v.agarwal@surrey.ac.uk; peter.young@kcl.ac.uk; sagar.joglekar@kcl.ac.uk; n.sastry@surrey.ac.uk.}
\email{v.agarwal@surrey.ac.uk}
\affiliation{%
  \institution{University of Surrey}
  \city{Guildford}
  \state{Surrey}
  \country{UK}
}

\author{Anthony P. Young}
\email{peter.young@kcl.ac.uk}
\affiliation{%
 \institution{King's College London}
 \city{London}
 \country{UK}
}

\author{Sagar Joglekar}
\email{sagar.joglekar@kcl.ac.uk}
\affiliation{%
 \institution{King's College London}
 \city{London}
 \country{UK}
}

\author{Nishanth Sastry}
\email{n.sastry@surrey.ac.uk}
\affiliation{%
  \institution{University of Surrey}
  \city{Guildford}
  \state{Surrey}
  \country{UK}
}

\renewcommand{\shortauthors}{Vibhor Agarwal, et al.}

\begin{abstract}
Online forums that allow \apy{for} participatory engagement between users have been transformative for \apy{the} public discussion of \apy{many} important issues. However, \apy{such} conversations can sometimes escalate into full-blown exchanges of hate and misinformation. Existing approaches in natural language processing (NLP), such as deep learning models for classification tasks, use \apy{as inputs} only a \apy{single} comment or a pair of comments depending upon \apy{whether the task concerns the inference of properties of the individual comments or the replies between pairs of comments, respectively}. But in online conversations, \apy{comments and} replies may be based on external context beyond \apy{the immediately relevant information that is input to the model. Therefore, being aware of the conversations' surrounding} contexts \apy{should improve the model's performance for the inference task at hand}.

We propose \textit{GraphNLI}\footnote{This paper is an extended version of \cite{agarwal2022graphnli} published in The ACM Web Conference 2022.}, a novel graph-based deep learning architecture that uses graph \apy{walks to incorporate} the wider context of a conversation in a principled \apy{manner}. Specifically, \apy{a graph walk starts from a given comment and samples ``nearby'' comments in the same or parallel conversation threads, which results in} additional embeddings \apy{that are aggregated together with the initial comment's embedding}. We then use these \apy{enriched} embeddings for downstream NLP prediction tasks that are important for online conversations. We evaluate \apy{GraphNLI} on two \apy{such} tasks - \textit{polarity prediction} and \apy{\textit{misogynistic hate speech detection}} - and found that our model consistently outperforms all \apy{relevant} baselines \apy{for both tasks}. Specifically, GraphNLI with a biased root-seeking random walk performs with a macro-$F_1$ score of $3$ and $6$ percentage points better than the best-performing BERT-based baselines for the polarity prediction and hate speech detection tasks, respectively. We also perform extensive ablative experiments and hyperparameter \apy{searches} to understand the efficacy of GraphNLI. This demonstrates the potential of context-aware models to capture the global context along with the local context of online conversations for these two tasks.


\end{abstract}



\maketitle

\section{Introduction}\label{sec:intro}

The Internet has \apy{empowered} people to \apy{take part} in sharing their views \apy{and debating} \apy{about many topics online}, often as \apy{written} comments or posts \apy{that \textit{argue} for some claim}. \apy{Such online} debates can \apy{often} become large and acrimonious\apy{, with some escalating into full-blown exchanges of hate and misinformation}. As many of these debates concern topics of societal importance\apy{, such as health and politics}, it is \apy{crucial} to be able to model these debates accurately and at scale, so that we can better understand and control \apy{for} phenomena such as the spread of hate~\cite{cinelli2021online,guest2021expert,jahan2021systematic,koffer2018discussing}, fake news~\cite{allcott2017social,hanselowski2018retrospective}, how best to moderate political polarisation \cite{bail2018exposure}, and how to break echo chambers by \apy{connecting} appropriate \apy{users who possess} opposing views \apy{about the same issue}~\cite{garimella2017reducing}.

An important task in modelling online debates is to be able to predict whether the reply of one comment to another is \textit{attacking} (disagreeing) or \textit{supporting} the \apy{comment} it is replying to. This relation of agreement (support) or disagreement (attack) of a reply is known as its \textit{polarity}~\cite{cayrol2005acceptability}\apy{, and we call the task of predicting this relation the \textit{polarity prediction task}}. The ability to accurately \apy{infer} the polarity of  replies in \apy{a large} online debate can allow us to measure \apy{various} properties of the debate, such as how ``controversial'' a discussion is, e.g. by counting the \apy{total} number of supporting 
\textit{vs.} attacking replies in the discussion~\cite{boschi2021has}, \apy{or model the ``controversy'' generated by each comment as the ratio of attacking to supporting replies to that comment}. Perhaps more importantly, if the polarity is known, we can then use techniques from \textit{argumentation theory}, a branch of artificial intelligence \apy{concerned} with the formal representation and resolution of disagreements \cite{rahwan2009argumentation}, to compute which arguments have been attacked and \apy{\textit{should} be} rebutted, and which ones stand unrebutted \apy{and thus \textit{should} be believed.}

Another important task in modelling online conversations is to infer whether the text of individual comments contain hate speech; we call this the \textit{hate speech detection task}. \textit{Hate speech}, informally defined, is public speech that incites negativity, hatred and even violence against an individual or groups of people for reasons solely based on perceived group-level and stereotypical attributes such as sex, sexual orientation, race, nationality, religion and political beliefs~\cite{jahan2021systematic}. Hate speech, if left unchallenged, can normalise unhealthy attitudes towards groups of people, divide societies and cause real harm to individuals~\cite{chetty2018hate}. Automatically and accurately identifying whether a piece of text contains hate speech at scale is highly non-trivial, given the many ways hate speech can and should be operationalised in a precise way while maintaining an awareness to the relevant historical, conceptual and ethical issues arising, in addition to the relative lack of good annotated data~\cite{guest2021expert,jahan2021systematic}. If a good model can be trained to accurately identify hate speech, then we can study effects such as how the proportion of comments posted to online debates that contain hate speech can change over time, or whether moderation policies can be successful at encouraging and targeting the refutation of hate speech through subsequent comments submitted.

Polarity prediction and hate speech detection are examples of common tasks that are approached by applying natural language processing (NLP) techniques. NLP models typically make their predictions based on the  natural language \apy{texts of  single comments, or at most, pairs of comments such as a reply and the post it is replying to.} \cite{cabrio2013natural,cocarascu2017identifying,guest2021expert}. 
However, by considering the \apy{input comment(s)} in isolation from the rest of the discussion, such approaches risk loosing crucial information. For example, in a large discussion thread with many comments, one comment  can easily be taken out of context and misunderstood as ``hate''. Similarly, if two correspondents have been replying to each other across multiple posts in a discussion (\eg{} user B replies to A and A then replies back to B), it is conceivable that an incorrect polarity may be inferred if one looks at only a reply and the immediate post it is replying to in isolation to the existing context.

In this paper, we call the input \apy{comment(s)} the \textit{local} context for the tasks and \apy{comments beyond those inputs, for example, those contained in the same or parallel discussion threads,} the \textit{global} context. We ask and answer the question: \textit{Can we improve the performance on both the polarity prediction and hate speech detection tasks by incorporating additional \apy{global} context beyond the most relevant comments \apy{in the local context}?}

Typically, \apy{an online discussion} can be seen as a \apy{directed} \textit{tree}, starting with an original post \apy{-} the \textit{root} of the tree \apy{-} and each reply $b$ \apy{creates} a directed edge \apy{to the node $a$ that it is replying to}.\footnote{\apy{In the rest of this paper we will use the terms ``comment'', ``post'', ``argument'' and ``node'' interchangeably. Further, we adopt the terminology that a ``child'' node points to (replies to) its ``parent'' node.}} \apy{The tree structure derives from the often-true property that every non-root comment can only reply to \textit{one} other comment.\footnote{\apy{We will briefly discuss non-tree debates in Section \ref{sec:conclusions}.}}} \apy{For the polarity prediction task, we} hypothesise that nodes \apy{``near''} $a$ and $b$, e.g. their \apy{descendants}, ancestors and siblings in the discussion tree, contain additional \apy{context} that may help understand whether $b$ is attacking \apy{or supporting} $a$. For example, if other siblings of $b$ (i.e. children of $a$ other than $b$) are also attacking $a$, then it \textit{may} be more likely that $b$ is also an attacking reply. \apy{Similarly, for the hate speech detection task, we hypothesise that understanding posts ``near'' $a$ can better predict whether $a$ contains hate speech or not, compared to just knowing $a$ itself.} \textit{Our key idea is to use \textit{graph walk techniques} to discover and utilise this neighbouring context in a principled fashion.} \apy{The contributions of this paper are as follows:}

\begin{enumerate}
\item We \apy{define random walks on the discussion trees} that sample \apy{additional, ``nearby'' nodes} of the global context in online discussions\apy{; these} additional nodes along with the local context\apy{, appropriately featurised and aggregated, will serve as the input to our model.}
\item We present \textbf{GraphNLI}~\footnote{``NLI'' stands for ``\textit{natural language inference}'' - see Section~\ref{sec:polarity_prediction_task}.} -- a novel graph-based deep learning architecture that is capable of accurately predicting reply polarity and accurately detecting misogynistic hate speech. We provide an open source implementation of the model for the community.\footnote{The model code and the dataset is available at \url{https://netsys.surrey.ac.uk/datasets/graphnli/}, last accessed 5 November 2022.}
\item We compare and contrast several NLP models, including Sentence-BERT~\cite{reimers-2019-sentence-bert}, to establish \apy{relevant baselines} for \apy{both} the polarity prediction \apy{and the misogynistic hate speech detection tasks. For the polarity prediction task, we will use data from Kialo\footnote{See \url{https://www.kialo.com/}, last accessed 5 November 2022.} that has been used in prior work \cite{agarwal2021under,boschi2021has,young2020ranking}. For the misogynistic hate speech detection task, we will use data from Reddit\footnote{See \url{https://www.reddit.com}, last accessed 5 November 2022.} that has been  labelled by experts~\cite{guest2021expert}. Both datasets are} in the form of discussion trees where the nodes are \apy{comments} submitted to the discussion and the edges denote which comments reply to which other comments.
\item \apy{After training GraphNLI, we find that our} model outperforms \apy{all of these} baselines in both the tasks. Specifically, GraphNLI with a weighted average, biased root-seeking random walk (see Section \ref{sec:graph-walks}) classifies polarity with an accuracy of $82.95\%$ and $78.96\%$ macro-$F_1$ on the test set, while the best baseline, Sentence-BERT, has an accuracy of $79.86\%$ and macro-$F_1$ of $75.81\%$. Further, GraphNLI with a weighted average, biased root-seeking random walk detects misogynistic hate speech with an accuracy of $93.18\%$ and macro-$F_1$ of $74.79\%$ on the test set, while the best baseline is BERT, which has an accuracy of $92.28\%$ and macro-$F_1$ of $68.79\%$. These results suggest that knowing the context of online discussions does help with the ability to classify whether a reply is supportive or attacking, and whether the text of a comment contains misogynistic hate speech.
\item We also systematically investigate through ablation studies what features can be helpful in capturing the wider context for both the tasks and show that upstream text, the parent and other ancestor nodes, help the model more than siblings and children replies. Moreover, we find that in the best-performing versions of GraphNLI, the importance of neighbouring nodes decreases as their distance from the given node increases. We also perform error analysis for the hate speech detection task and show that GraphNLI gives less numbers of false positives and false negatives due to the context-awareness of online conversations.
\end{enumerate}


The rest of this paper is structured as follows. \apy{Section \ref{sec:background}} provides an overview of \apy{both} the polarity prediction \apy{and hate speech detection tasks. In Section \ref{sec:methodology}, we will define and classify various kinds of graph walks by the probability of the walk going ``up'' the tree towards the root and how the contextual nodes captured are discounted. We will also define the architecture for GraphNLI. In Section \ref{sec:applications}, we overview the Kialo and Reddit datasets used to train and evaluate GraphNLI. In Section \ref{sec:results}, we train and evaluate different kinds of GraphNLI based on their graph walks against various baselines for both tasks, and show that GraphNLI outperforms all baselines. We will also discuss hyperparameter searches and conduct} an ablation study to better understand which features are important. In Section~\ref{sec:discussion}, we perform error analysis\apy{, and give an example from the Reddit dataset where hateful speech, that has been misidentified as non-hate by BERT, is correctly identified by GraphNLI given the latter's awareness of the conversation context, as well as non-hate that has been misidentified as hate by BERT is correctly identified as non-hate by GraphNLI. We conclude and outline possible future works in Section \ref{sec:conclusions}.}

\section{Background}\label{sec:background}

\subsection{Polarity Prediction}\label{sec:polarity_prediction_task}

Suppose we have an online debate that has the structure of a directed tree, where the set of nodes $A$ denote the arguments submitted to the debate, and the directed edges $E\subseteq A\times A$ denote which arguments reply to which other arguments; the tree structure means that, apart from the very first argument submitted in the debate, each argument replies to exactly one other argument. The \textit{polarity prediction task} asks: for the argument $b$ that is replying to the argument $a$, is this reply in \textit{agreement} (\textit{support}) or \textit{disagreement} (\textit{attack})? Note that both of these categories are defined commonsensically and not rigorously, such that (e.g.) an attack does not have to contain some logical contradiction.

\begin{example}\label{eg:vegan}
Let $a$ be the argument or thesis that ``All humans should be vegan.'' Let $b$ be the argument, ``Veganism is a restrictive diet and is not a healthy lifestyle for many people''. Further, let $b$ reply to $a$.\footnote{This example is taken from \url{https://www.kialo.com/all-humans-should-be-vegan-2762}, last accessed 5 November 2022.} 
Most English-speaking people who understand the statements of $a$ and $b$ should agree that argument $b$ \textit{disagrees} with argument $a$. Can a machine learning model replicate such predictions accurately and at scale?
\end{example}

The polarity prediction task is one of the many tasks in the field of \textit{argument mining} (e.g. \cite{lawrence2020argument,lippi2016argumentation,cabrio2018five}); this is the application of NLP techniques to extract arguments and identify their relationships from raw text. Other example tasks include identifying when Tweets from Twitter are well-defined arguments instead of insults, single URLs or pictures \cite{bosc2016tweeties}, identifying the claims, their reasons and the relationships between these claims from clinical trials to inform medical decision making \cite{mayer2021enhancing}, or detecting fallacies from the transcripts of the United States Presidential Debates~\cite{villata2021argstrength}. The polarity prediction task is important because once we have classified all replies in a debate as either supporting or attacking, we can apply ideas from argumentation theory to reason about which arguments should be justified given the information presented. \textit{Argumentation theory} is a branch of AI that is concerned with the transparent and rational resolution of disagreements (e.g. \cite{rahwan2009argumentation}). Formally, online debates as described above can be represented as a \textit{bipolar argumentation framework} (BAF) (e.g. \cite{cayrol2005acceptability}), which is a triple $\ang{A,R_{att},R_{sup}}$ where $A$ is the set of arguments submitted in the debate, $R_{att}\subseteq A^2$ is the set of edges that are attacking, and $R_{sup}\subseteq A^2$ is the set of edges that are supporting. It is required that $R_{sup}\cap R_{att}=\emptyset$. Resolving the disagreements formally amounts to selecting a subset of arguments $S\subseteq A$ that satisfy various normative criteria (e.g. \cite{baroni2011introduction,dung1995acceptability,young2018approx}). For example, the property of \textit{conflict-freeness}, $(S\times S)\cap R_{att}=\emptyset$, formalises the idea of self-consistency because winning arguments should not attack each other.

\begin{example}\label{eg:vegan2}
(Example \ref{eg:vegan} continued) Let the argument $c$ be, ``It is often necessary for a vegan diet to \href{https://www.healthline.com/nutrition/7-supplements-for-vegans}{include supplements}, as it lacks specific \href{https://www.healthline.com/nutrition/7-nutrients-you-cant-get-from-plants#4.-Vitamin-D3-(cholecalciferol)}{essential nutrients}''.\footnote{From \url{https://www.kialo.com/it-is-often-necessary-for-a-vegan-diet-to-include-supplements-as-it-lacks-specific-essential-nutrients-2762.502?path=2762.0~2762.1-2762.11731_2762.502}, last accessed 5 November 2022.} Let the argument $d$ be, ``\href{https://pubmed.ncbi.nlm.nih.gov/26502280/}{Research} results suggest that no significant vitamins or minerals deficiencies affect the vegan population compared to non vegans''.\footnote{From \url{https://www.kialo.com/research-results-suggest-that-no-significant-vitamins-or-minerals-deficiencies-affect-the-vegan-population-compared-to-2762.1066?path=2762.0~2762.1-2762.11731_2762.502-2762.1066}, last accessed 5 November 2022.} We have the argument $c$ \textit{supporting} the argument $b$ (from Example \ref{eg:vegan}), and the argument $d$ \textit{attacking} the argument $c$. Suppose that these arguments are all there is of interest, then the corresponding BAF is $A=\set{a,b,c,d}$, $R_{sup}=\set{(c,b)}$ and $R_{att}=\set{(b,a),(d,c)}$. We can visualise this BAF in Figure~\ref{fig:BAF}.

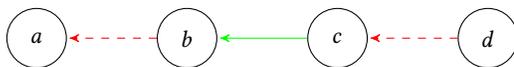
\begin{figure}[h]
\begin{center}
\begin{tikzpicture}[>=stealth',shorten >=1pt,node distance=2cm,on grid,initial/.style    ={}]
\tikzset{mystyle/.style={->,relative=false,in=0,out=0}};

\node[state] (a) at (0,0) {$ a $};
\node[state] (b) at (2,0) {$ b $};
\node[state] (c) at (4,0) {$ c $};
\node[state] (d) at (6,0) {$ d $};
\draw [->, red, dashed] (b) to (a);
\draw [->, red, dashed] (d) to (c);
\draw [->, green] (c) to (b);

\end{tikzpicture}
\caption{The bipolar argumentation framework in Example \ref{eg:vegan2} visualised. The broken (red) arrows denote attacks, and the solid (green) arrows denote supports.}\label{fig:BAF}
\end{center}
\end{figure}
\end{example}

Many other kinds of analyses can be performed once a debate has been represented as a BAF. For instance, based on the polarities of the edges, we may calculate which arguments should be justified and which arguments have been rebutted. This could potentially be used to present only the justified arguments as summary to a reader. Previous work has also looked at how the conclusions of an idealised reader can change depending on  which parts of a debate they read, thus underscoring the dangers of sampling only parts of a large online debate~\cite{young2018approx,young2020ranking,young2021likes}. Other work has shown how the location of the justified arguments can be significantly influenced by whether the debate is acrimonious or supporting~\cite{boschi2021has,young2022modelling}. BAFs are therefore useful as they allow for the application of both argument-theoretic and graph-theoretic ideas to gain insights about online discussions \cite{young2022modelling}.

The polarity prediction task has been discussed in the argument mining literature. For example, \cite{cabrio2018five} has reviewed the task in the context of persuasive essays or political debates. An early example of this work is \cite{cabrio2013natural}, which applied textual entailment (e.g. \cite{bos2006logical,dagan2010recognizing,kouylekov2010open,maccartney2008modeling}) to predict the polarity of replies on the now-defunct Debatepedia dataset,\footnote{This is archived in \url{http://web.archive.org/web/20201008080532/http://www.debatepedia.org/en/index.php/Welcome_to_Debatepedia\%21}, last accessed 5 November 2022.} with a test accuracy of $67\%$. \textit{Textual entailment} is the task of identifying the relationship between an ordered pair of texts, specifically whether the first entails the second, or contradicts the second, or is neutral, and is also called \textit{natural language inference}.\footnote{See, e.g. \url{https://nlp.stanford.edu/projects/snli/}, last accessed 4 November 2022; our framework is named \textit{GraphNLI} as it was first designed for polarity prediction, which is conceptually similar to textual entailment.} In \cite{cocarascu2017identifying}, long-short-term memory networks were used to classify polarity, achieving $89\%$ accuracy.\footnote{Although this shows better results on the polarity prediction task than what we report in Section \ref{sec:results}, neither their data nor their framework \apy{were} available for benchmarking.} A more recent overview of the polarity prediction task \cite{cocarascu2020dataset} has provided context-independent baselines of neural network models using a range of learning representations and architectures, and have found an averaged performance of $51\%$ to $55\%$ of these different neural networks across such contexts; these contexts involve online debates on a range of controversial topics such as abortion and gun rights, persuasive essays, and presidential debates. 

In all of the above-mentioned approaches, the inputs to the model are the texts of the replying argument and the argument being replied to, often represented by some appropriate word embedding. Arguably, this is the least amount of information one must input into the model to predict the polarity of the reply. What has not yet been considered is whether it is helpful to incorporate \textit{more} information. \apy{Naively, one can input the \textit{entire} conversation network into the model, although this is easily intractable when the sizes of such networks are large. But suppose we were to add more context in an \textit{incremental} fashion.} For example, if argument $c$ replies to argument $b$, and $b$ replies to argument $a$, and we would like to predict the polarity of the reply from $c$ to $b$, then is it useful to design a model that accepts as input the texts of $c$, $b$ \textit{and} $a$? How about if we randomly\footnote{``Randomly'' in a well-defined sense, see Sections \ref{sec:graph-walks} and \ref{sec:graph-walks2}.} sample additional ``nearby'' comments to build up a context? To the best of our knowledge, these questions have not yet been addressed in the argument mining literature. We thus seek to investigate these questions by measuring whether models that incorporate additional context in online conversations outperform models that are not aware of this extra information on the polarity prediction task.

\subsection{Hate Speech Detection}\label{sec:hate_speech_detection}

As mentioned in Section \ref{sec:intro}, Internet debates, especially those about controversial topics, can easily spread hate and misinformation. One reason is that many people, with a range of personalities and behavioural dispositions, can easily access the Internet and participate in various online forums or microblogging services; the relative anonymity the Internet offers can encourage some people to behave in ways that are unacceptable in the offline world, often resulting in the spread of hate speech \cite{cinelli2021online,guest2021expert,jahan2021systematic,koffer2018discussing}.

Hate speech is notoriously difficult to define. A sample of important attempted definitions (e.g. \cite[Section 2.1]{jahan2021systematic}) all agree that hate speech is public language that attacks individuals and groups of people because of protected characteristics, for example, their race, skin colour, religion, ancestry, nationality, gender, disability, sexuality and so on. Hate speech, if left unchallenged, can promote and incite harmful societal consequences against individuals and groups such as (but not limited to) physical attacks, psychological intimidation, property damage, violence and segregation. Therefore, it is important to \textit{at least} be able to detect hate speech in online forums, accurately and at scale, such that appropriate action can be taken by the moderators, which can range from banning people who continuously send unambiguously hateful messages, or encouraging more moderate users on how to publicly refute such arguments in a civil and transparent manner.

How can we detect hate speech in a scalable and accurate manner? If there exists an expertly-annotated textual dataset that clearly denotes and explains which texts are examples of hate speech, then we can use it to train various NLP models to classify and even explain whether a piece of text contains hate or not. However, this brings us back to our initial problem - how should hate speech be defined in a manner that is sensitive to the relevant historical, conceptual and ethical issues that arise, and is untainted by ideology of the kind that seeks to silence opposing viewpoints by deliberately taking things out of context? Further, how should such a definition inform the labelling of data, especially data from online discussions?


\begin{example}\label{eg:hate_speech_context}
Consider the Reddit conversation in Figure \ref{fig:guest-fn-eg}.

\begin{figure}[h]
\centering
\includegraphics[width=0.6\linewidth]{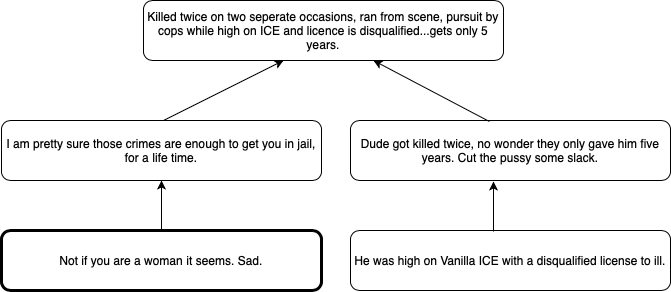}
\caption{A fragment of a Reddit conversation from the dataset in \cite{guest2021expert}.}
\label{fig:guest-fn-eg}
\end{figure}

\noindent The rectangular text boxes represent posts submitted to this particular Reddit discussion, and the arrows denote which posts reply to which other posts. Suppose we wish to identify whether the text with the thicker border in the bottom left contains hate speech. This text states, ``\textit{Not if you are a woman it seems. Sad.}'' At first glance, this text appears to mention something about women and as such does not appear to be hate speech. Indeed, it could even seem like a post sympathetic to the difficulties or biases that women may encounter. However, considering the context of the surrounding comments in the thread, it is clear that the post is derogatory to women, claiming that women are not adequately sentenced like men and therefore an instance of misogyny. Most hate speech and misogyny models are currently unable to capture this nuance without any conversational context.
\end{example}


Examples like this demonstrate that although hate exists and should be dealt with accordingly, the accurate detection of hate speech is very important as speech misunderstood as hateful or not hateful can have real consequences in the offline world. Further, the real risk of people taking things out of context motivates the application of a similar technique to improving the polarity prediction task as discussed in Section \ref{sec:polarity_prediction_task}: can hate speech models be improved by systematically incorporating the surrounding context in online conversations?

To begin to answer this question, we build on the recent work of Guest et al. \cite{guest2021expert}, which is concerned with misogynistic speech in Reddit. Their contributions include a high-quality, expert-labelled dataset from various Reddit communities based on a clear taxonomy of misogynistic speech \cite[Section 4]{guest2021expert}; each of these labels have been checked by an average of three annotators. Further, three baseline classifiers were offered - logistic regression, unweighted BERT and weighted BERT, with respective $F_1$ scores of $0.13$, $0.42$ and $0.43$ \cite[Section 7 and Appendix C]{guest2021expert} when classifying individual Reddit posts for misogyny on the basis of their text. This work therefore provides a clean dataset of misogynistic speech, and several classifier baselines on which one can improve on. 

Like Kialo, Reddit discussions have an explicit reply tree structure; this allows us to incorporate context by incrementally and systematically including more ``nearby'' comments by aggregating their word embeddings with the embedding of the node to be classified as whether it contains misogynistic speech or not. Does this incorporation of additional context help misogynistic hate speech models better detect online hate speech?

\section{Methodology}\label{sec:methodology}


\apy{As stated in Sections \ref{sec:intro} and \ref{sec:background}, various} deep learning models have been used in the literature to perform \apy{NLP} classification tasks \apy{concerning online conversations}. However, depending on the task and model, most approaches usually consider either a \apy{single} comment \apy{to be classified}, or take a pair of comments such as a reply and the post it is replying to. In this section, we propose a novel graph-based deep learning architecture that not only considers as input the \apy{single} comment \apy{or the pair of comments}, but also systematically captures the context of nearby comments via graph walks.

\subsection{Representing Online Discussions as Trees}\label{sec:discussion-trees}

For every online discussion \textit{D}, we construct a \apy{discussion tree}, where a node represents a post / comment / argument and \apy{the edges are \textit{directed} from a given node to the other unique node it is replying to. \apy{The discussion forms a tree structure because it starts with a \textit{root} node (out-degree = 0), and every non-root node replies to exactly one other node (out-degree $=1$), while all nodes can have zero or more replies to it (in-degree $\in\nat$).}} Each such node has an associated label depending upon the prediction task. For polarity prediction, \apy{the non-root nodes are labelled with} \textit{support} or \textit{attack}, depending upon whether the \apy{post} is respectively for or against its parent \apy{post. For misogynistic} hate speech, \apy{each node} is \apy{labelled as} either \textit{hate} or \textit{non-hate}. The root node of this discussion tree represents the \apy{opening comment or post} of the discussion. \apy{Every non-root node replies to exactly one other node in the tree.}

\subsection{GraphNLI Architecture}

\apy{In this subsection, we define different kinds of graph walks and explain how they (probabilistically) sample neighbouring nodes to feed in the global context into our classifier. Further, distant nodes can have their influences discounted by some appropriate discount factor (as a multiplicative factor to their word embeddings). We then provide the architecture of GraphNLI, and explain how it makes inferences given the input discussion node (pairs) and their surrounding context as sampled from the graph walks.}

\subsubsection{\textbf{Capturing Global Context through Graph Walks}}
\label{sec:graph-walks}

GraphNLI captures the global context of online conversations through graph-based walks. A \textit{walk} is defined as a \apy{finite} sequence of nodes traversed from a given node in a tree\apy{, such that each adjacent pair in the sequence is joined by an edge. In Section \ref{sec:discussion-trees}, we have stated that our trees are directed, such that the edge directions denote which comments reply to which other comments. However, the walks we consider}  \textit{\textit{ignore} the direction of the edges}\apy{; traversal \textit{against} the edge directions allows the capture of} context such as\apy{ the} children or siblings (reachable by going up to the parent node and then \textit{back down again} to the sibling) of the reply node. We define $L$ to be the maximum number of distinct nodes sampled by a graph walk including the starting node, then the walk length is $(L - 1)$.

\apy{To sample the neighbouring nodes, we propose a \textit{biased root-seeking random walk}}.
In a discussion tree, a \textit{biased root-seeking random walk} is a walk that starts from a given node and traverses other nodes probabilistically, but it is \textit{biased} towards the root. This bias captures the intuition that the ``best'' or most relevant context might be found in the sub-thread of replies leading from the root down to the node we are considering. 

\apy{Recall that each non-root node replies to \textit{exactly one} other node, and each node has a non-negative number of nodes replying to it. Suppose we begin the walk on a non-root node $a_1$. Let $p\in[0,1]$ be the probability that the next node traversed in the walk is the unique node that $a_1$ replies to.}
To bias the walk towards the root node, we \apy{choose} $p>=0.5$. \apy{As} we ignore the direction of edges in these walks, \apy{this allows} the walk to traverse \textit{downwards} as well as upwards in the tree. The remaining probability, $1-p < 0.5$, is divided equally among all the \apy{children nodes replying to $a_1$. We illustrate this with Example \ref{eg:walk1}.}

\begin{example}\label{eg:walk1}
Suppose we have a section of a discussion tree with comments $a_0$, $a_1$, $a_2$, $a_3$ and $a_4$ and edges $(a_1,a_0)$, $(a_2,a_1)$, $(a_3,a_1)$ and $(a_4,a_1)$, such that \apy{$(a_i,a_j)$ means $a_i$ replies to $a_j$}. Let $p=0.75$, then $1-p=0.25$. A \apy{biased root-seeking} random walk starting at $a_1$ will have probability $0.75$ moving \textit{up} to $a_0$ next. Similarly, starting from $a_1$, there is a  probability \apy{$\frac{0.25}{3}=\frac{1}{12}$} of moving \textit{down} (against the directions of the arrows) to any of $a_2$, $a_3$, or $a_4$. \apy{This is shown} in Figure~\ref{fig:random-walk-ex3.1}.

\begin{figure}[h]
\centering
\includegraphics[width=0.4\linewidth]{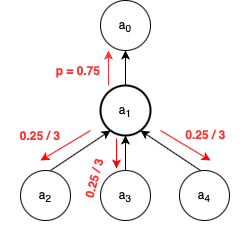}
\caption{The possible next node for a biased root-seeking random walk with $p=0.75$, starting at $a_1$, from Example \ref{eg:walk1}.}
\label{fig:random-walk-ex3.1}
\end{figure}
\end{example}

\apy{Such biased root-seeking random walks therefore sample $L$ nodes in the discussion tree to be inputted into the GraphNLI model. However, as the walk is random and can go against the direction of reply, it is possible that the same node is visited more than once (this is consistent with the use of the term ``walk'' in graph theory, as opposed to ``path'' where vertices cannot repeat). If this happens, we ignore any duplicate nodes and continue the walk until either $L$ \textit{distinct} nodes are traversed, or the walk} terminates when no more distinct nodes are available to be sampled. \apy{We illustrate this with Example \ref{eg:walk2}.}

\begin{example}\label{eg:walk2}
Suppose we have a section of a discussion tree with nodes $a_0$, $a_1$, $a_2$, $a_3$ and $a_4$ and edges $(a_1,a_0)$, $(a_2,a_1)$, $(a_3,a_1)$ and $(a_4,a_2)$, such that \apy{$(a_i,a_j)$ denotes that $a_i$ replies to $a_j$.} Let $p=0.75$, then $1-p=0.25$. \apy{A random walk, starting at $a_2$, can move to $a_1$ with probability $0.75$, or to $a_4$ with probability $0.25$. Suppose that the walk moves to $a_1$, then there is a probability of $0.25$ that the walk will move \textit{down} and sample $a_2$ again. But since $a_2$ has already been visited,} it will be ignored and the walk continues. \apy{This is shown in Figure~\ref{fig:random-walk-ex3.2}.}

\begin{figure}[h]
\centering
\includegraphics[width=0.3\linewidth]{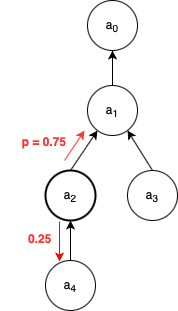}
\caption{The possible next node for a biased root-seeking random walk with $p=0.75$, starting at $a_2$, from Example \ref{eg:walk2}.}
\label{fig:random-walk-ex3.2}
\end{figure}
\end{example}

\apy{Notice that as $p\in[0,1]$, it is possible to choose $p=1$, which results in a \textit{deterministic} walk towards the root node of the discussion.} Agarwal et al.~\cite{agarwal2022graphnli} \apy{calls this a \textit{root-seeking graph walk; it} is clearly} a special case of biased root-seeking random walk. \apy{Intuitively, this walk captures \textit{only} the prior context of the discussion leading up to the node of interest.}

It is important to limit the walk length as online conversations can grow rapidly, and capturing far away nodes through \apy{such} random walks can lead to \apy{the} over-smoothing problem~\cite{chen2020measuring}; \apy{the} phenomenon in which the resulting node embeddings of different pieces of text are almost indistinguishable and consequently, the model looses its \apy{predictive} power \apy{due to} capturing \textit{too much} context. The walk length $L-1$ determines the maximum length of this random walk, where $L$ is the maximum number of distinct nodes to be visited from the starting node until the random walk terminates. By experimenting with walk length $L$ and evaluating how well GraphNLI performs, we found $L = 4$ to be the optimal walk length (see Section \ref{sec:results}).

\begin{example}
(Example \ref{eg:walk2} continued) Consider a root-seeking graph walk starting at $a_2$, where by definition $p=1$. Consider sampling $L=4$ arguments. The result would be $(a_2,a_1,a_0)$ and then nothing else, as the walk can no longer move ``up'' after reaching the root, $a_0$. Therefore, $L$ is an upper bound on the number of nodes sampled.
\end{example}

The \apy{biased root-seeking} random walk is \apy{thus} a parametrised way of randomly sampling neighbouring nodes as a means of incorporating the \apy{global} context for the prediction tasks. Note that there is no guarantee that the parent node will be visited in the random walk. By choosing the \apy{value of} $p$, we can directly affect the probability of visiting the parent as discussed in Section~\ref{sec:ablation-study}. Even if the parent is not visited, there is likely to be information in the surrounding nodes that still helps in the prediction task. For example, if the majority of children nodes replying to the parent are attacking (perhaps because the parent post is controversial), knowing the sibling context may help in the prediction task.

\subsubsection{\textbf{Discounting the Influence of Distant Nodes}}\label{sec:graph-walks2}

\apy{Once we have at most $L$ distinct nodes sampled using the biased root-seeking random walk, we can obtain each node's text and thus each text's corresponding word embedding vector. We discount} the contributions of \apy{each} neighbouring node's corresponding embedding vector by a \apy{weight} of $\gamma^k$, where $\gamma\in[0,1]$ is the \textit{discount parameter} and $k$ is the distance along the graph walk from the starting node. This gives a \textit{weighted} random walk, \apy{where the highest weight $\gamma^0$ is given to the starting node, the second-highest weight $\gamma^1$} is given to the immediate neighbour (either parent or a child node), then a discounted weight to the neighbour's neighbour in a random walk, and so on. This means closer the sampled node is to the starting node along the graph walk, the higher its weight will be, and the more it\apy{s embedding vector} will contribute as an input towards the prediction task \apy{concerning the starting node}.

\begin{example}
(Example \ref{eg:walk2} continued) Suppose that $L=4$ and the resulting random walk starting at $a_2$ gives $\pair{a_2,a_4,a_1,a_0}$. Suppose $\gamma=0.5$. Let $v_i$ be the embedding vector of the text of $a_i$, for $i\in\set{0,1,2,4}$. The resulting weighted vectors are $v_2$, $\gamma v_4$, $\gamma^2 v_1$ and $\gamma^3 v_0$. Notice that the powers of $\gamma$ refer to the position along the graph walk that a new node is encountered, and not to the graph-theoretic distance between nodes. For example, node $a_1$ is one edge away from $a_2$, yet its discount factor is $\gamma^2$, not $\gamma^1$ since it is at the second position in the graph walk starting from $a_2$.
\end{example}

\noindent If $\gamma=1$, then all the nodes sampled by the walk will have equal influence, regardless of their distance from the starting node. Conversely, if $\gamma=0$ and by adopting the convention that $\lim_{x\to 0^+}x^x=1$,\footnote{This limit can be informally verified by, e.g. plotting the graph of $y=x^x$ for $x\in\real^+$, and is a convention widely adopted in mathematics, e.g. in information theory when calculating entropy.} then all the nodes will have zero weight apart from the starting node. Intuitively, $\gamma$ is thus a measure of how much the model should care about the surrounding context.

At the end of a graph walk for each node, we obtain at most \textit{L} distinct comments which \apy{includes the starting comment and its ancestors, descendents and siblings}. We input these sets of comments into our GraphNLI model.

\subsubsection{\textbf{Model Overview}}\label{sec:model_overview}

\begin{figure}[h]
\centering
\includegraphics[width=0.6\linewidth]{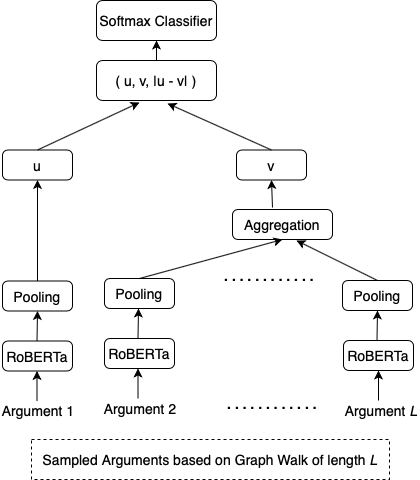}
\caption{\apy{The architecture for GraphNLI - suppose we want to classify whether the argumentative text in node $a$ contains misogynistic hate speech, or to classify whether the reply from $a$ to node $b$ is an attack or a support between arguments (assuming such a reply exists, but if it exists it is unique). We use a graph walk (Section~\ref{sec:graph-walks}) to sample up to $L$ arguments for classifying a single node; we assume there are exactly $L$ such arguments sampled in the diagram for simplicity. Each argument is represented by its RoBERTa embedding, and then mean pooling is applied to obtain sentence embeddings. The arguments beyond the first are aggregated to an embedding vector $v$; we will compare the various aggregation methods in Section \ref{sec:ablation-study}. The first argument is represented with a vector $u$. This is then concatenated into an embedding vector three times as long: $(u,v,|u-v|)$ where the last one denotes the element-wise difference of the embeddings of $u$ and $v$. This aggregated representation serves as the input to a softmax classifier for the downstream classification task.}}\label{fig:graphnli-arch}
\end{figure}

GraphNLI is a novel graph-based deep learning architecture which captures both the local and the global context of the online conversations through graph-based walks, \apy{as explained in the preceding two subsections}. \apy{The architecture for GraphNLI is shown in Figure \ref{fig:graphnli-arch}.}


\apy{GraphNLI} is inspired by S-BERT~\cite{reimers-2019-sentence-bert}. Firstly, each of the arguments sampled by the graph walk, of which there is at most $L\in\nat^+$ (Section \ref{sec:graph-walks}), is input into the RoBERTa model~\cite{liu2019roberta} to get their corresponding embeddings. \apy{Then,} a mean-pooling operation, that is, \apy{calculating the} mean of all the output vectors, is applied to derive a fixed-sized sentence embedding for each argument. The starting node in a graph walk is a \textit{point-of-interest} (PoI) node. Let $u$ denote the sentence embedding corresponding to the PoI node. Let $v$ denote the aggregated embedding from its \apy{contextual nodes' text embeddings} (\apy{which may or may not include the} parent) sampled by the graph walk starting from the PoI node. These $u$ and $v$ embeddings together are then used to predict the polarity \apy{of the reply from the PoI node to its parent, or whether the PoI node's text contains misogynistic hate speech}. We \apy{have} experiment\apy{ed} with three aggregation strategies: \apy{\textit{summation} (component-wise sum), \textit{average} (component-wise arithmetic mean) and \textit{weighted average}} to compute the resultant embedding $v$. As stated in Section~\ref{sec:graph-walks2}, \apy{the} nodes sampled by a root-seeking random walk are weighted \apy{by powers of $\gamma$} in descending order from the PoI node up to the root. \apy{Given $u$ and $v$, we calculate an element-wise difference vector $|u-v|$. We then concatenate all three vectors $u$, $v$ and $|u-v|$ together} to get the final embedding vector, which is then fed into a softmax classifier for the downstream prediction task.

In order to fine-tune BERT, we make the GraphNLI model end-to-end trainable to update weights during backpropagation such that the \apy{resulting} sentence embeddings are semantically meaningful for the \apy{various} downstream prediction tasks.

\section{\apy{Prediction Tasks and Datasets}}\label{sec:applications}

\subsection{\apy{The Polarity Prediction Task and the Kialo Dataset}}\label{sec:kialo_dataset}

As discussed in Section \ref{sec:polarity_prediction_task}, polarity prediction aims to identify the argumentative relations of \textit{attack} and \textit{support} between natural language arguments, where in our case such \apy{arguments} are comments submitted to online debates, and one text is replying to another text~\cite{agarwal2022graphnli}.

\apy{We use a dataset from Kialo to train GraphNLI.} \textit{Kialo} is an online debating platform that helps people ``engage in thoughtful discussion, understand different points of view, and help with collaborative decision-making''.\footnote{Quoted from \url{https://www.kialo.com/about}, last accessed 2 November 2022.} In this study, we use data from discussions hosted on the Kialo debating platform as used by \cite{agarwal2022graphnli,boschi2021has,young2020ranking,young2022modelling}. In a Kialo debate, users submit \textit{claims} \apy{supported by reasons, so each claim is an argument. Each claim submitted after the very first claim of a debate replies to exactly one other claim. The first claim submitted in a debate is its \textit{thesis}, which does not reply to anything - this means Kialo debates are trees and the thesis argument is the root of the tree}. To start a discussion in Kialo, the user creates a thesis along with a tag that indexes the discussion by indicating the content of the discussion. A thesis can have many tags, which increases its visibility to the users. Users then comment on the discussions of their choice. The dataset contains $1,560$ discussion threads \apy{dated until} 28 January 2020. Each discussion thread has data about the tree structure, votes on each argument's impact on the debate it has been submitted to, and the arguments' texts. Further, each reply between arguments is clearly labelled as attacking (negative) or supporting (positive). Table~\ref{tab:dataset-stats} below shows the \apy{number of examples per class}. On each \apy{discussion tree}, there is a reasonable amount of debate, with a mean of $204$ and a median of $68$ arguments (standard deviation $463$). Kialo debates are typically balanced, with the vast majority of discussion trees having around $40\%$ of its replies as supporting, with the rest being attacking (see Table \ref{tab:dataset-stats}).

Due to Kialo's strict moderation policy, each piece of text submitted to a debate is a self-contained argument that has clear claim backed by reasons. Thus, each post in Kialo can be taken as a node and directed edges can be drawn based on which post is replying to which other post. The polarity prediction task is to decide whether these edges are attacking or supporting.

\subsection{The \apy{Misogynistic} Hate Speech Detection Task and the Reddit Dataset}\label{sec:guest_dataset}

As discussed in Section \ref{sec:hate_speech_detection}, hate speech is prevalent in social media \apy{and it is important to at least be able to accurately identify its occurrence in a scalable manner, such that appropriate action can be taken. The hate speech detection task} aims to predict whether a comment is \textit{hate} or \textit{non-hate} in online conversations.

\apy{To evaluate how well GraphNLI classifies text as hate speech by being aware of the context of the discussion surrounding each piece of text, we use the dataset curated by Guest et al. \cite{guest2021expert}. This} is an expert-annotated, hate speech dataset, sourced from Reddit. This dataset looks at the specific type of hate against women - misogyny. Therefore, the positive class is ``misogynistic'' and negative class is ``non-misogynistic''. Each instance is annotated by three annotators on average, \apy{following a clearly-defined taxonomy that articulates various subtypes of misogynistic speech. However, for our purposes, we will only consider the most coarse-grained class - whether the speech is misogynistic or not.} Table~\ref{tab:dataset-stats} shows the \apy{number of examples per class}; we can see that misogynistic instances are \apy{in the} minority.

We specifically choose this dataset for hate speech detection because it has information needed to construct discussion trees from Reddit conversations\apy{, from which we can sample the appropriate context using graph walks as discussed in Section \ref{sec:graph-walks}. Specifically,} every \apy{post} has a parent \apy{ID, which allows us to infer the reply structure of the discussion tree. To the best of our knowledge, the baselines in \cite{guest2021expert} have not exploited the context.}

\begin{table}[h]
\begin{tabular}{c|ccccc}
\hline
\textbf{Task}                  & \textbf{Dataset} & \textbf{Positive class} & \textbf{Negative class} & \textbf{Positive class} $\%$ & \textbf{Negative class} $\%$ \\ \hline
Polarity prediction   & Kialo   & $139,722$      & $184,651$      & $43.1$              & $56.9$              \\
Hate speech detection & Reddit  & $699$          & $5,868$        & $10.6$              & $89.4$              \\ \hline
\end{tabular}
\caption{\apy{Class frequencies and percentages for each dataset. ``Positive'' refers to \textit{supportive} replies in Kialo, and the \textit{presence} of misogynistic hate speech in the Guest dataset. ``Negative'' refers to \textit{attacking} replies in Kialo, and the \textit{absence} of misogynistic hate speech.}}\label{tab:dataset-stats}
\end{table}



\section{Experiments and Results}\label{sec:results}

\subsection{Dataset Preprocessing}

As described in Section~\ref{sec:kialo_dataset}, we use data from online debates conducted on Kialo for polarity prediction. All discussions from Kialo have a tree structure with a root node that represents the main thesis and each other node is a reply to its parent, which either supports or attacks the parent. As discussed in \apy{Section~\ref{sec:graph-walks}, the graph walks treat Kialo debates as undirected discussion trees.} Each edge is either a support or attack. We randomly sample $80\%$ of the Kialo debates into a training set with the remainder serving as a test set. Overall, the training set contains $259,499$ arguments (replies) in total, while the test set contains $64,874$ arguments in Kialo debates.

For \apy{the} hate speech detection task, we use \apy{the} Guest dataset \apy{of misogynistic hate speech} as described in Section~\ref{sec:guest_dataset}. We represent Reddit conversations in the form of discussion trees and then we randomly sample $80\%$ of the conversations into train set and the rest $20\%$ into test set. The training set contains $5,335$ \apy{posts}, while the test set contains $1,516$ \apy{posts} in total.

\subsection{Training Details}

After preprocessing \apy{both} datasets, we use \apy{the various} graph walk techniques described in Section~\ref{sec:graph-walks} to capture the neighbourhood and parent contexts for each of the nodes and feed them into our GraphNLI model. In the case of \apy{the} biased root-seeking random walk, we \apy{perform} a detailed hyperparameter search on $p$ and $\gamma$ values\apy{, which we will describe} in Section~\ref{sec:hyperparameter-search}.


We fine-tune GraphNLI model with a softmax classifier objective function and cross-entropy loss for four epochs. We use a batch-size of 16, Adam optimizer with learning rate $2\times 10^{-5}$, and a linear learning rate warm-up over $10\%$ of the training data.

\subsection{Baselines}\label{sec:baselines}

We compare GraphNLI with the following relevant baselines. For \apy{the} polarity prediction task, we concatenate the embeddings of parent and child comments for the model input as the model needs to predict whether the child comment attacks or supports the parent comment. In case of \apy{the} hate speech detection task, we directly input the comment embeddings into the machine learning model.

\textbf{Bag-of-Words with Logistic Regression}: \apy{The first baseline is a} bag-of-words (BoW) model which uses unigram features as input obtained from the comments in online conversations. \apy{For the polarity prediction task, the inputs are the concatenations of} the parent and child BoW embeddings. \apy{For the hate speech detection task, the inputs are the single} comment BoW embeddings. \apy{The inputs with their corresponding labels for each task are then} fed into a \apy{logistic regression classifier with L2 regularization}. \apy{The classifier is} trained for $100$ epochs.


\textbf{BERT}: Bidirectional encoder representations from transformers (BERT)~\cite{devlin2019bert} is a transformer network~\cite{vaswani2017attention} pre-trained with a vast amount of raw text. It is one of the top performing models on various hate speech detection datasets \cite{jahan2021systematic}. \apy{As a baseline, it} is fine-tuned on the Guest training set for 4 epochs with a batch-size of 16, Adam optimizer, and cross-entropy loss function.

\textbf{Sentence-BERT}: S-BERT~\cite{reimers-2019-sentence-bert} is a modification of a pre-trained BERT transformer network to derive semantically meaningful sentence embeddings. We use the S-BERT architecture with a binary classification objective function and input the sentence pairs (parent and child arguments) into the model to get their sentence embeddings for \apy{the} polarity prediction task. Later on, these embeddings are concatenated and fed into a softmax classifier. The S-BERT model is fine-tuned on Kialo training dataset for 4 epochs with a batch-size of 16, the Adam optimizer, and \apy{binary} cross-entropy loss function.

\textbf{Non-trainable BERT embeddings with graph walks and Multi-layer Perceptron}: For each of the arguments in \apy{the two datasets}, their embeddings are derived using a pre-trained BERT model and using CLS-token embeddings. Using \apy{the various} graph walk techniques as described in Section~\ref{sec:graph-walks}, various neighborhood siblings and parent nodes are sampled for each node, and using their node embeddings, a resulting aggregated embedding is formed using an average aggregation function. These node embeddings are then fed into a multi-layer perceptron (MLP) with two layers and a softmax objective function for prediction. The initial BERT embeddings are non-trainable. We train the MLP for 50 epochs or until the model converges on Kialo training dataset with batch-size of 16, \apy{using} the Adam optimizer.

\subsection{Evaluation Metrics}\label{sec:evaluation-metrics}

Given the imbalanced nature of Guest dataset, we use the following metrics to evaluate our model and other baselines. Although Kialo dataset is balanced, we report all the metrics along with accuracy for consistency.

\begin{itemize}
\item \textbf{Accuracy}: Accuracy is the most intuitive performance measure and is a ratio of correctly predicted observations to the total observations.
\item \textbf{Macro-$F_1$}: Macro $F_1$ is the arithmetic mean of \apy{the $F_1$ scores per class.}
\item \textbf{Precision}: Precision is the ratio of correctly predicted positive observations to the total predicted positive observations.
\item \textbf{Recall}: Recall is the ratio of correctly predicted positive observations to the all observations in positive class.
\end{itemize}

\noindent \apy{Accuracy and macro-$F_1$ provide a} high-level representation of the overall model performance. \apy{Precision and recall are used} to evaluate the model's ability to predict the minority class.

\subsection{Model Evaluation}

\subsubsection{\textbf{Performance on the Kialo Dataset}}\label{sec:evaluation1}

For \apy{the} polarity prediction task, we evaluate the performance of GraphNLI on the test set of Kialo data. We use the various evaluation metrics as discussed in Section~\ref{sec:evaluation-metrics} to \apy{verify} the model effectiveness. We train models with five different random seeds and report their average performances. Table~\ref{tab:results} shows the performance of \apy{the various} models \apy{on the polarity prediction task after being} trained on \apy{the same} Kialo \apy{training} set.

\begin{table}[h]
\begin{tabular}{l|cccc}
\hline
\textbf{Model} & \textbf{Accuracy} & \textbf{Macro-$F_1$} & \textbf{Precision} & \textbf{Recall} \\
\hline
Bag-of-Words + Logistic Regression & 67.00 & 62.00 & 62.00 & 62.00  \\
Sentence-BERT with classification layer & 79.86 & 75.81 & 77.86 & 73.86     \\
BERT Embeddings: Root-seeking Graph Walk + MLP & 70.27 & 52.32 & 44.87 & 64.12    \\
\hline
GraphNLI: Root-seeking Graph Walk + Sum & 80.70 & 77.97 & 78.02 & 77.93     \\
GraphNLI: Root-seeking Graph Walk + Avg. & 81.86 & 77.98 & 78.17 & 77.84    \\
GraphNLI: Root-seeking Graph Walk + Weighted Avg. & 81.97 & 76.89 & 76.83 & 76.96  \\
GraphNLI: Biased Root-seeking Random Walk + Sum & 79.95 & 76.32 & 76.29 & 76.35     \\
GraphNLI: Biased Root-seeking Random Walk + Avg. & 80.44 & 76.62 & 76.68 & 76.56    \\
GraphNLI: Biased Root-seeking Random Walk + Weighted Avg. & \textbf{81.95} & \textbf{78.96} & \textbf{78.94} & \textbf{78.99}   \\
\hline
\end{tabular}
\caption{Performance on \textit{Kialo} dataset for polarity prediction, discussed in Section \ref{sec:evaluation1}. In case of root-seeking graph walk, $p=1$ and $\gamma=0.8$, whereas in root-seeking random walk, $p=0.8$ and $\gamma=0.8$.}\label{tab:results}
\end{table}

The baseline model, Bag-of-Words embeddings with Logistic Regression, achieves an accuracy of $67\%$ with $62\%$ macro-$F_1$ score. The most relevant and state-of-the-art Sentence-BERT model trained on Kialo dataset achieves an accuracy of $79.86\%$ and macro-$F_1$ of $75.81\%$. The initial MLP model with non-trainable BERT embeddings and root-seeking graph walk achieves an accuracy of $70.27\%$ which is even worse than the Sentence-BERT.

Our model, GraphNLI, with the root-seeking graph walk (probability $p = 1$) and averaging node embeddings in the graph walk to get the aggregated node embeddings achieves an overall accuracy of $81.86\%$, whereas, the model achieves even better accuracy of $81.97\%$ using weighted average node embeddings with macro-$F_1$ score of $76.89\%$. In the case of the biased root-seeking random walk and averaged node embeddings, GraphNLI achieves an accuracy of $80.44\%$ with $76.62\%$ macro-$F_1$. The best performing version of GraphNLI is with biased root-seeking random walk and weighted average aggregation, obtaining $81.95\%$ accuracy and $78.96\%$ macro-$F_1$. This implies a significant improvement of about 4 percentage points in macro-$F_1$ and 3 percentage points in the accuracy score. Clearly, all the variants of GraphNLI achieve better accuracy scores than all the baselines, including sentence-BERT. GraphNLI with a biased root-seeking random walk and weighted average node embeddings with probability $p = 0.8$ and gamma $\gamma = 0.8$ achieves the highest accuracy and macro-$F_1$ overall. We will discuss detailed hyperparameter search in Section~\ref{sec:hyperparameter-search}. The results show that systematically incorporating the global context of the online debates or discussions along with the local context of the argument pairs does help in predicting the argumentative relations of support and attack. Also, weighted average aggregation gives higher weights to the arguments near to the given argument pair in the discussion tree whose polarity needs to be predicted, and exponentially reduces the weights when the graph walk moves away from the given node.

\subsubsection{\textbf{Performance on the Guest Dataset}}\label{sec:evaluation2}

We evaluate the performance of GraphNLI model on the Reddit dataset by Guest et al. \cite{guest2021expert} for the misogynistic hate speech detection task. Table~\ref{tab:perf-eacl} shows the performance of different models on the test set.

\begin{table}[h]
\begin{tabular}{l|cccc}
\hline
\textbf{Model} & \textbf{Accuracy} & \textbf{Macro-$F_1$} & \textbf{Precision} & \textbf{Recall} \\
\hline
Bag-of-Words + Logistic Regression & 92.08 & 61.45 & 56.98 & 71.49  \\
BERT with classification layer & 92.28 & 68.79 & 74.96 & 65.47 \\
BERT Embeddings: Root-seeking Graph Walk + MLP & 92.14 & 66.56 & 61.89 & 71.65  \\
\hline
GraphNLI: Root-seeking Graph Walk + Sum & 92.63 & 72.74 & 78.34 & 69.33 \\
GraphNLI: Root-seeking Graph Walk + Avg. & 92.95 & 72.91 & 78.46 & 69.92 \\
GraphNLI: Root-seeking Graph Walk + Weighted Avg. & 93.06 & 73.56 & 80.63 & 69.48    \\
GraphNLI: Biased Root-seeking Random Walk + Sum & 93.05 & 74.34 & 80.37 & 70.26 \\
GraphNLI: Biased Root-seeking Random Walk + Avg. & 93.16 & 74.56 & 80.24 & 70.81 \\
GraphNLI: Biased Root-seeking Random Walk + Weighted Avg. & \textbf{93.18} & \textbf{74.79} & \textbf{80.89} & \textbf{70.90}    \\
\hline
\end{tabular}
\caption{Performance on \textit{Guest} dataset for hate speech detection, discussed in Section \ref{sec:evaluation2}. In case of root-seeking graph walk, $p=1$ and $\gamma=0.2$, whereas in root-seeking random walk, $p=0.6$ and $\gamma=0.2$.
}\label{tab:perf-eacl}
\end{table}

We have trained models with five different random seeds and report their average performances. The dataset is highly imbalanced\apy{, with the hate class in the minority,} hence we use different evaluation metrics such as macro-$F_1$, precision, and recall as described in Section~\ref{sec:evaluation-metrics}. The baseline model, Bag-of-Words embeddings with Logistic Regression, achieves an overall accuracy of $92.08\%$ with $61.45\%$ macro-$F_1$ score. The state-of-the-art BERT model achieves an overall accuracy of $92.28\%$, $68.79\%$ macro-$F_1$, $74.96\%$ precision, and a poor recall of $65.47\%$. The initial MLP model with non-trainable BERT embeddings and root-seeking graph walk achieves an accuracy of $92.14\%$ with $66.56\%$ macro-$F_1$. GraphNLI with a root-seeking graph walk (probability $p = 1$) performs the best with weighted average aggregation\apy{; this has} an overall accuracy of $93.06\%$ and macro-$F_1$ score of $73.56\%$. All the different versions of GraphNLI outperform the baselines on all evaluation metrics with a significant improvement in macro-$F_1$. The best performing variant is GraphNLI with a biased root-seeking random walk and weighted average aggregation with probability $p = 0.6$ and gamma $\gamma = 0.2$. We will discuss the hyperparameter search in detail in Section~\ref{sec:hyperparameter-search}.

The best performing model gives an overall accuracy of $93.18\%$ and macro-$F_1$ of $74.79\%$. This implies a significant improvement in precision, recall, and macro-$F_1$, which is about $6$ percentage points higher than the BERT baseline. Once again, this shows that systematically incorporating the global context of the conversations along with the local context improves the model performance in detecting hate speech online.

\subsection{Hyperparameter Search and Ablative Experiments}\label{sec:hyperparameter-search}

\subsubsection{\textbf{Hyperparameter Search}}

In this section, we \apy{comment on the performance of GraphNLI on both tasks given the values of various hyperparameters, and how such values were searched for}.

\apy{Recall from Section \ref{sec:graph-walks} that in} a biased root-seeking random walk, \apy{the} probability $p$ is the probability of selecting the parent node \apy{(i.e. the node being replied to by the current, point-of-interest (PoI) node in a discussion tree)}, while the remaining probability $1 - p$ is distributed equally among the children nodes \apy{of the PoI node}. Recall from Section \ref{sec:graph-walks2} that gamma $\gamma$ is the discount parameter which exponentially decreases as the random walk moves away from the \apy{PoI} node towards one of the parent or children nodes. \apy{For both the polarity prediction and the misogynistic hate speech detection tasks, we} perform \apy{a} detailed hyperparameter search on different pairs of values of $p$ and $\gamma$ ranging from $0$ to $1$ \apy{in increments of $0.2$, and observe their macro-$F_1$ scores}.

\begin{figure}[h]
\centering
\includegraphics[width=0.6\linewidth]{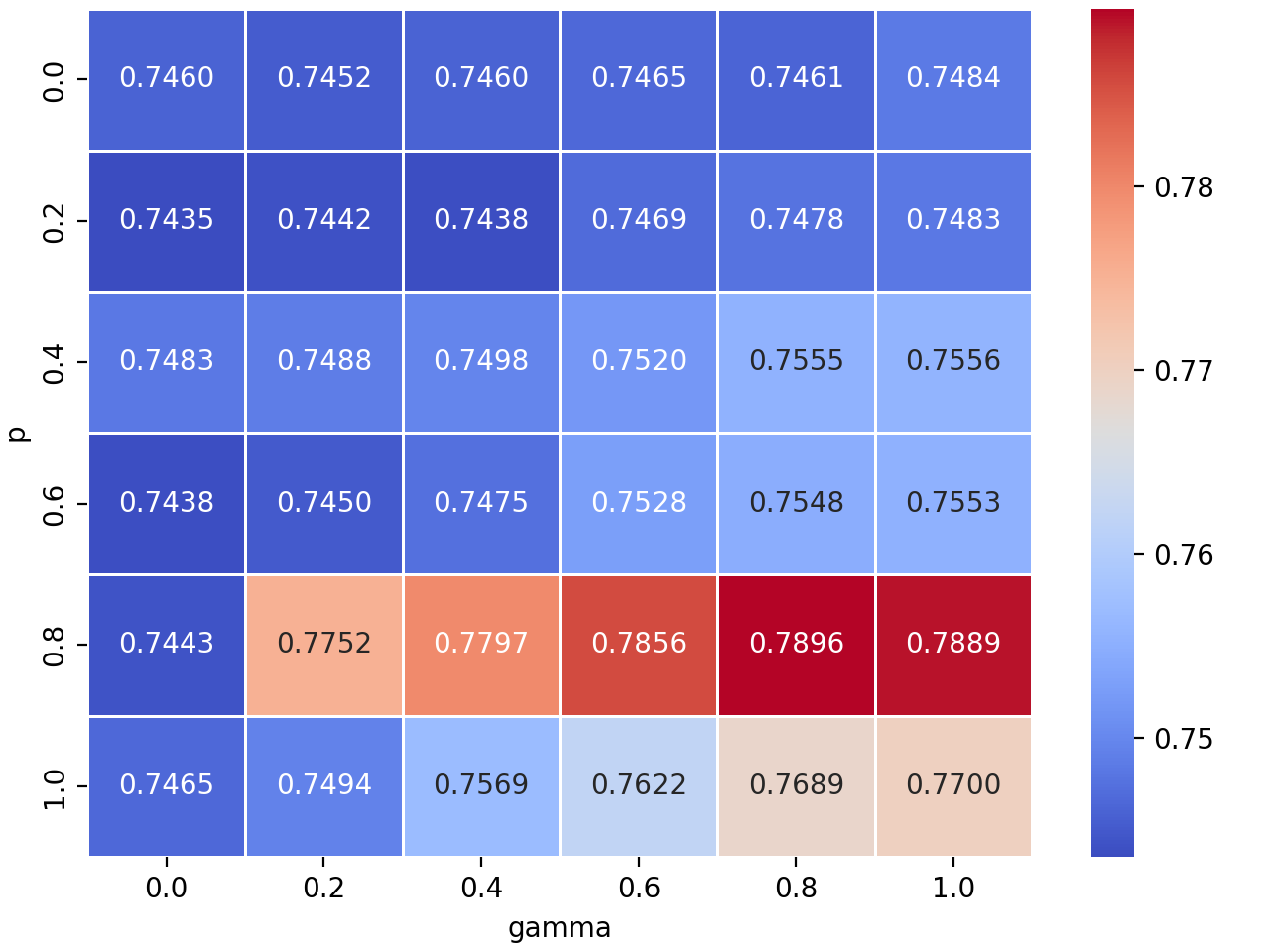}
\caption{\apy{A heat map showing the macro-$F_1$ scores for different values of gamma ($\gamma$) and probability $p$ on the polarity prediction task, where the GraphNLI model has been trained on the Kialo dataset by Agarwal et al.~\cite{agarwal2022graphnli} }}\label{fig:kialo-f1-heatmap}
\end{figure}

Figure~\ref{fig:kialo-f1-heatmap} shows the heat map of \apy{the} macro-$F_1$ scores for GraphNLI \apy{when} trained on Kialo dataset for different values of \apy{the discount factor} $\gamma$ and probability $p$. \apy{If $p<0.5$, then the} children nodes \apy{of the PoI node} will be selected \apy{via} random walk more often than the parent and other ancestor nodes\apy{. Conversely, if $p> 0.5$, then the} parent and ancestor nodes \apy{of the PoI node} will be selected more often. For a root-seeking random walk, \apy{the} macro-$F_1$ score of $78.96\%$ is the highest when $p = 0.8$ and $\gamma = 0.8$. \apy{A h}igher value of \apy{the} probability ($p = 0.8$) shows that the importance of parent and other ancestor nodes is more than the children nodes for \apy{the} polarity prediction task, whereas \apy{a} higher value of gamma ($\gamma = 0.8$) indicates that immediate neighbors of a node (argument) matter more than the nodes which are farther away. For a root-seeking graph walk ($p = 1.0$), macro-$F_1$ is the highest when $\gamma = 1.0$ (all the nodes in a graph walk are weighted equally), but still lower than the root-seeking random walk. This demonstrates the importance of capturing and weighting the neighbouring context along with the local context for polarity prediction.

\begin{figure}[h]
\centering
\includegraphics[width=0.6\linewidth]{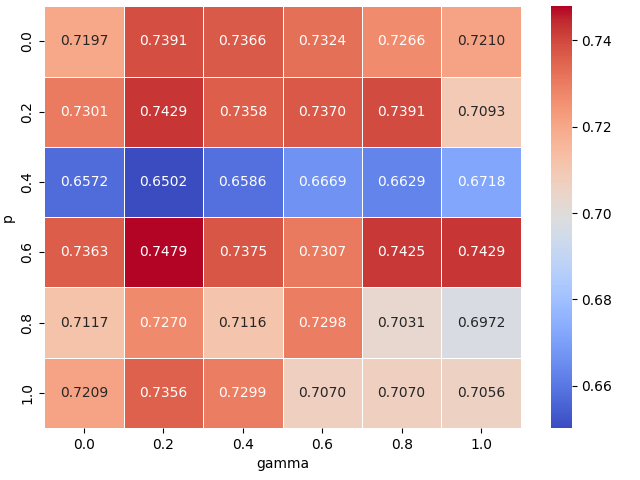}
\caption{\apy{A heat map showing the macro-$F_1$ scores for different values of gamma ($\gamma$) and probability $p$ on the misogynistic hate speech detection task, where the GraphNLI model has been trained on the Reddit dataset by Guest et al. \cite{guest2021expert}}.}\label{fig:eacl-f1-heatmap}
\end{figure}

\apy{For the hate speech detection task, the analogous hyperparameter search results with the same ranges for $\gamma$ and $p$ are shown in the heat map in Figure~\ref{fig:eacl-f1-heatmap}.}

For a root-seeking random walk, a macro-$F_1$ score of $74.79\%$ for hate speech detection is the highest when $p = 0.6$ and $\gamma = 0.2$. \apy{As $p=0.6>0.5$,} ancestor nodes will be selected more often by such a random walk than the children nodes. A low value of $\gamma$, i.e. $0.2$, means that context of nodes which are farther away matters a lot along with the immediate neighbours. For a specific random walk -- a root-seeking graph walk ($p = 1$), the highest macro-$F_1$ score of $73.56\%$ \apy{occurs} when $\gamma = 0.2$. It is $1$ percentage point lower than the highest macro-$F_1$ for a biased root-seeking random walk. This demonstrates the importance of capturing the context of children and other sibling nodes along with the ancestor nodes for hate speech detection. Both Figures \ref{fig:kialo-f1-heatmap} and \ref{fig:eacl-f1-heatmap} show that inputting neighbouring context is always better than no context.

\subsubsection{\textbf{Ablative Experiments}}\label{sec:ablation-study}

We have demonstrated superior performance of GraphNLI with respect to \apy{the} various baselines in Table~\ref{tab:results} for polarity prediction and \apy{in} Table~\ref{tab:perf-eacl} for hate speech detection. In this section, we perform ablative experiments and discuss different aspects of the GraphNLI model \apy{as well as the} intuition behind the choices in order to gain a better understanding of the model.

First, we evaluate different graph walks by feeding the resultant embeddings obtained with the weighted average aggregation strategy into our model and compare their accuracy and $F_1$-scores. For both the tasks, as shown in Tables~\ref{tab:results} and \ref{tab:perf-eacl}, all the graph walks with \apy{the} GraphNLI architecture perform \apy{significantly} better than the BERT-based model. S-BERT just considers the argument pairs (node and its parent embeddings) for polarity prediction and BERT just considers the node embedding for hate speech detection, whereas through graph walks, GraphNLI considers the global context of the discussion trees by exploring parents and neighborhoods of a node. Therefore, the global context along with the local context of the online discussions indeed helps in predicting the polarities \apy{of replies and the presence of misogynistic hate speech}.

We further evaluate different aggregation strategies (summation, average and weighted average) to aggregate the node embeddings of the neighbouring nodes using \apy{the biased root-seeking random walk}. As shown in Tables~\ref{tab:results} and \ref{tab:perf-eacl}, the weighted average aggregation function performs better than the summation and average strategies. This shows that influence of the neighbouring comments decreases as the graph walk moves away from the given node. Hence, neighbouring nodes cannot be weighted equally but instead, progressively in the decreasing order of their distance from the given node.

\begin{table}[h]
\begin{tabular}{lcc}
\toprule
\textbf{Concatenation} & \textbf{Kialo} & \textbf{Guest} \\
\midrule
$(u, v)$ & 76.78 & 92.71 \\
$(u, v, u*v)$ & 82.05 & 93.06 \\
$(u, v, |u-v|)$ & \textbf{82.87} & \textbf{93.18} \\
$(u, v, |u-v|, u*v)$ & 82.38 & 92.48 \\
\bottomrule
\end{tabular}
\caption{Accuracy (\%) scores of GraphNLI model trained on \apy{the Kialo and Guest datasets} with different concatenation techniques using weighted average aggregation, discussed in Section \ref{sec:ablation-study}. \apy{The concatenations are denoted by tuples of the vectors concatenated in order, where $u*v$ denotes element-wise multiplication and $|u-v|$ denotes element-wise modulus of the differences.}}\label{tab:ablation-results}
\end{table}

We also evaluate different  methods for concatenating a node's embedding \textit{u} with the aggregated embedding of its neighbours \textit{v} obtained using the root-seeking random walk. The impact of the concatenation method on the model's performance is significant. As depicted in Table~\ref{tab:ablation-results}, the concatenation of $(u, v, |u-v|)$ works the best for both the tasks. As reported by~\cite{reimers-2019-sentence-bert}, adding the element-wise multiplication $u * v$ decreased the performance. Element-wise absolute difference $|u - v|$ which measures the distance between the two node embeddings is \apy{thus} an important component.

Our initial non-end-to-end trainable model as discussed in Section~\ref{sec:baselines} in which we keep the node embeddings obtained from the BERT model fixed, performs even worse than Sentence-BERT. This throws light on the importance of end-to-end training of the model for fine-tuning on specific tasks. After end-to-end training, the model outputs node embeddings that are rich in context suitable for downstream tasks like polarity prediction and hate speech detection.

\section{Error Analysis on Hate Speech Detection}\label{sec:discussion}

In this section, we conduct an error analysis on hate speech detection and compare our GraphNLI model with the best performing baseline -- BERT for hate speech detection. Table~\ref{tab:hate-p-r-f1} shows the precision, recall, and $F_1$-score for the \textit{hate} class in Guest dataset. The best performing variant of GraphNLI with biased root-seeking random walk ($p = 0.6$ and $\gamma = 0.2$) is compared with the best performing baseline BERT on hate class. The BERT baseline gives an overall hate $F_1$-score of $45\%$ with hate precision of $53\%$ and a poor hate recall of $39\%$. GraphNLI gives a $7$ percentage point higher hate $F_1$-score with an overall hate precision of $65\%$ and hate recall of $44\%$. The improvement in hate precision is the highest at $12$ percentage points and hate recall at $5$ percentage points. Therefore, GraphNLI performs significantly better than BERT in detecting hate speech in online conversations.

\begin{table}[h]
\begin{tabular}{lccc}
\toprule
& \textbf{Hate Precision} & \textbf{Hate Recall} & \textbf{Hate $F_1$} \\
\midrule
BERT & 0.53 & 0.39 & 0.45 \\
GraphNLI & 0.65 & 0.44 & 0.52 \\
\bottomrule
\end{tabular}
\caption{Hate precision, recall, and $F_1$-scores for models trained on the Guest dataset.}\label{tab:hate-p-r-f1}
\end{table}


\apy{We look at the confusion matrix in} Table~\ref{tab:confusion-matrix-bert} for \apy{the baseline} BERT model trained on Guest dataset. Overall, 98 samples ($7.96\%$) are misclassified. Out of these 98 samples, 36 ($36.74\%$) are false positives and 62 ($63.27\%$) are false negatives.

\begin{table}[h]
\begin{tabular}{ll|ll|l}
\hline
& & \textbf{Prediction} & \\
& & Non-hate & Hate & Total  \\
\hline
\textbf{Label} & Non-hate &  1,093 & 36 & 1,129 \\
& Hate & 62 & 40 & 102 \\
\hline
& Total & 1,155 & 76 & 1,231 \\
\hline
\end{tabular}
\caption{Confusion matrix of the BERT model trained on \textit{Guest} dataset.}\label{tab:confusion-matrix-bert}
\end{table}


\noindent Table~\ref{tab:confusion-matrix-graphnli} shows the confusion matrix for GraphNLI model trained on Guest dataset. Overall, 82 samples ($6.66\%$) are misclassified, which is less than the BERT baseline. Out of these 82 samples, 28 ($34\%$) are false positives and 54 ($66\%$) are false negatives.

\begin{table}[ht]
\begin{tabular}{ll|ll|l}
\hline
& & \textbf{Prediction} & \\
& & Non-hate & Hate & Total  \\
\hline
\textbf{Label} & Non-hate & 1,101 & 28 & 1,129 \\
& Hate & 54 & 48 & 102 \\
\hline
& Total & 1,155 & 76 & 1,231 \\
\hline
\end{tabular}
\caption{Confusion matrix of the GraphNLI model trained on \textit{Guest} dataset.}
\label{tab:confusion-matrix-graphnli}
\end{table}

\apy{Our conclusion is that \textit{GraphNLI's ability to incorporate the global context of online conversations leads to lesser false positives and false negatives compared to the current-best performing model that does not incorporate context.}} Out of 62 false negatives for BERT model, many cases are misclassified due to the lack of context of conversation threads. Thus, the BERT model, which does not explicitly take into account the conversation context, fails to recognise the misogyny in isolation.

\begin{example}
We now consider a false negative example from the \textit{Guest} dataset, as shown in Figure~\ref{fig:repeated_guest_example}. This is the same conversation from Example \pageref{eg:hate_speech_context}, but we repeat it here for convenience.

\begin{figure}[ht]
\centering
\includegraphics[width=0.6\linewidth]{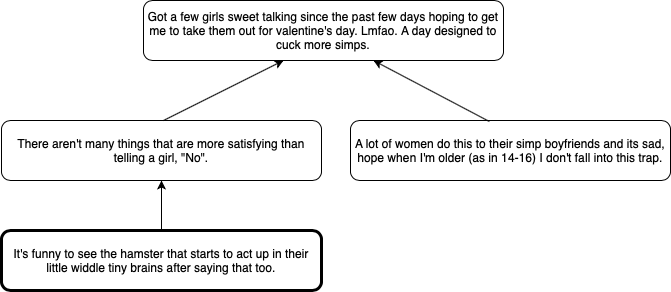}
\caption{An example conversation from \textit{Guest} dataset containing a false negative sample.}\label{fig:repeated_guest_example}
\end{figure}

\noindent The nodes of this directed tree are Reddit posts, whose texts are:

\begin{enumerate}
\item \textbf{Bottom row:} ``It's funny to see the hamster that starts to act up in their little widdle tiny brains after saying that too''. This is labelled as misogynistic as the annotators agree that the text points out that women are intellectually inferior.
\item \textbf{Middle row, left:} ``There aren't many things that are more satisfying than telling a girl, `No'.''
\item \textbf{Middle row, right:} ``A lot of women do this to their simp boyfriends and its sad, hope when I'm older (as in 14-16) I don't fall into this trap.''
\item \textbf{Top row:} ``Got a few girls sweet talking since the past few days hoping to get me to take them out for valentine's day. Lmfao. A day designed to cuck more simps.''
\end{enumerate}

\noindent The arrows between the nodes indicate which posts reply to which other posts.

We would like to automatically detect whether the post on the bottom row contains misogynistic hate speech. But BERT cannot recognise that as it does not have enough context to understand that it refers to women. On the other hand, GraphNLI successfully recognises the bottom post as misogynistic since it has additional context of the surrounding comments, some of which are shown in Figure \ref{fig:guest-fn-eg}. In this manner, GraphNLI gives $8$ lesser false negatives than BERT.
\end{example}

We now look at a false positive example.

\begin{example}
Consider a false positive example from the \textit{Guest} dataset, as shown in Figure~\ref{fig:guest-fp-eg}.

\begin{figure}[h]
\centering
\includegraphics[width=0.6\linewidth]{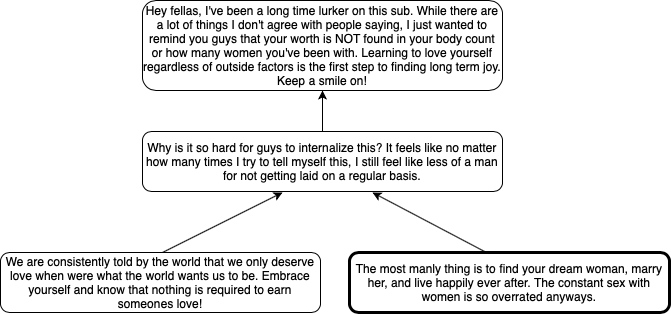}
\caption{An example conversation from \textit{Guest} dataset containing a false positive sample.}
\label{fig:guest-fp-eg}
\end{figure}

\noindent The nodes of this directed tree are Reddit posts, whose texts are:

\begin{enumerate}
\item \textbf{Bottom row, left:} ``We are consistently told by the world that we only deserve love when were what the world wants us to be. Embrace yourself and know that nothing is required to earn someones love!''
\item \textbf{Bottom row, right:} ``The most manly thing is to find your dream woman, marry her, and live happily ever after. The constant sex with women is so overrated anyways.'' This is labelled as non-misogynistic by expert annotators. Although text contains few words that can overlap with misogynistic content, overall it suggests that things other than high levels of sexual activity should be prioritised, and therefore it is labelled as non-misogynistic.
\item \textbf{Middle row:} ``Why is it so hard for guys to internalize this? It feels like no matter how many times I try to tell myself this, I still feel like less of a man for not getting laid on a regular basis.''
\item \textbf{Top row:} ``Hey fellas, I've been a long time lurker on this sub. While there are a lot of things I don't agree with people saying, I just wanted to remind you guys that your worth is NOT found in your body count or how many women you've been with. Learning to love yourself regardless of outside factors is the first step to finding long term joy. Keep a smile on!''
\end{enumerate}

We would like to automatically detect whether the post on the bottom right row contains misogynistic hate speech. Although it is labelled as non-misogynistic, BERT recognises it as misogynistic as the post contains few words often associated with misogynistic content. On the other hand, GraphNLI successfully recognises it as non-misogynistic since it has additional context of the surrounding comments, some of which are shown in Figure~\ref{fig:guest-fp-eg}.
\end{example}

\noindent Overall, due to context awareness, GraphNLI gives $8$ lesser false positives compared to BERT.

\section{Conclusions and Future Work}\label{sec:conclusions}

In this paper, we demonstrated a novel model, GraphNLI, which is capable of classifying the polarity of replies (attacking or supporting) in an online discussion, as well as whether the texts posted in such discussions contain misogynistic hate speech. The classification component of GraphNLI is inspired by S-BERT, but it is novel in its application of graph walks to sample the ``global'' context surrounding the reply or post in an online discussion, and then using this context to enrich its inputs. Empirically, we found that a biased root-seeking random walk with a weighted average aggregation of the neighbouring contexts is the best strategy in terms of classification accuracy. This strategy addresses the shortcomings of previous approaches that only capture the ``local'' context, such as the reply and the post it is replying to. We also showed through extensive ablative experiments that information from parent and other ancestor nodes provide more relevant contextual information than siblings and children of the reply node for both the classification tasks. Furthermore, the importance of ancestor nodes should decrease as the distance from the reply node increases. The ability for GraphNLI to outperform the current state-of-the-art for two tasks shows the value of incorporating the context of online discussions.

Our work partly fixes the gap of inferring polarity and hate speech from the content and context of an online discourse. However, several questions remain to be answered to make it relevant for wider usage. For instance, we have demonstrated and evaluated our approach on Kialo, a tightly moderated online debating platform, as well as noisier, weakly moderated discourses, such as on Reddit. However, other forums, such as BBC's \textit{Have Your Say}?\footnote{\apy{See the comments of, e.g. \url{https://www.bbc.co.uk/news/uk-politics-63484971\#comments}, last accessed 2 November 2022.}} do not have an explicit threaded reply structure, requiring one to \textit{infer} from the text of a reply which other post it is replying to, prior to applying GraphNLI's graph walk techniques. In such less restrictive user interfaces, posts may refer to \textit{multiple} other posts, which in turn means that the reply graph is no longer a tree, although still a directed acyclic graph due to the time ordering of replies, i.e., that later posts can only reply to earlier posts. However, we note that discussions that are not trees should pose no limitations to our graph walk techniques (Section \ref{sec:graph-walks}) because there would only be more context for the graph walk to sample; future work will make this claim precise.

Further, GraphNLI can be applied to a range of problems in computational social science, for example:

\begin{itemize}
\item \textbf{Using reply polarity to enhance hate speech detection:} As stated in Section \ref{sec:hate_speech_detection}, hate speech on online forums is a  common challenge \cite{cinelli2021online}, including in nationally important conversations between citizens and their elected representatives~\cite{pushkal21MPHate}. In other cases, some members of a discussion can be unfairly targeted, as in the case of misogyny on Reddit~\cite{guest2021expert}. Knowledge about argument polarities can support the detection of hate speech, for example, many attacking comments towards female participants (if the gender of users are accurately known) can be a possible complementary indicator of potential misogyny.
\item \textbf{Understanding conversation health:} As stated in Section \ref{sec:intro}, online discussion forums provide a great opportunity for socially positive interactions, such as peer-support for long-term medical problems~\cite{joglekar2018online,panzarasa2020social}. On the other hand, many forums have unfortunately become a medium for rampant misinformation~\cite{kumar2016disinformation} and hate~\cite{gagliardone2015countering}. As such, identifying and promoting ``healthy'' conversations has been identified as an important priority by many (e.g. Twitter~\cite{twitter2022healthyconv}). Intuitively, an online discussion is ``healthier'' if there is less hate speech than non-hate speech, and that there is less attacks than there is support, adjusted for how controversial the topic under discussion is. This can be used to develop conversation health ``metrics'' which can be tracked over time, and guide appropriate action to be taken when necessary.
\item \textbf{Detecting filter bubbles:} Democratic conversations on news and social media sites can exhibit partisan tendencies~\cite{NRSWWW2018b,agarwal2021under,karamshuk16slant,NRSWWW2020}. This can lead to filter bubbles, where two (or more) parallel conversations about the same topic exist, with each conversation consisting of posts largely agreeing with other posts in that conversation, and yet having a large amount of disagreement with the other conversations happening in parallel. Predicting polarities could help detect filter bubbles by quantifying agreeability in conversations, e.g. if we find that posts reachable from each other also agree with each other (\textit{i.e.,} are supporting), and yet if an imaginary edge is induced between posts in different parts of a conversation (or a different discussion thread), we find that the imaginary edge would be an attack edge, this could be indicative of a filter bubble. Further, one can extend the ability of detecting hate speech to identifying the target of such hatred, which can complement such efforts in identifying filter bubbles.


\item \textbf{Eristic argumentation:} Formal models of argumentation (e.g. \cite{baroni2011introduction,cayrol2005acceptability,dung1995acceptability,young2018notes}) have focussed on logical and dialectical aspects of argumentation, such as formalising the central concepts of validity, justification and explainability. However, such formal models have until recently paid less attention to \textit{eristic argumentation} - where people argue for the sake of winning and \textit{causing} conflict, rather than \textit{resolving} conflict; such a style of argumentation is done without necessarily adhering to the norms of logic, facts or civil debate. Perhaps unsurprisingly, online debates are highly eristic, and there have been attempts to model this using formal argumentation (e.g. \cite{blount2014towards}). GraphNLI can contribute to this effort by relating how the presence of hate relates to when an online discussion that is initially civilised shifts to an eristic discussion, which would then suggest that more traditional models of argumentation may no longer apply.
\end{itemize}

We therefore hope to apply and adapt GraphNLI to some of the above problems in future work.

\bibliographystyle{ACM-Reference-Format}
\bibliography{sample-base}










\end{document}